\documentclass{article}

\usepackage[preprint]{arxiv}

\usepackage{bbm}
\usepackage[utf8]{inputenc} % allow utf-8 input
\usepackage[T1]{fontenc}    % use 8-bit T1 fonts
\usepackage{hyperref}       % hyperlinks

\usepackage{url}            % simple URL typesetting
\usepackage{booktabs}       % professional-quality tables
\usepackage{amsfonts}       % blackboard math symbols
\usepackage{nicefrac}       % compact symbols for 1/2, etc.
\usepackage{subcaption}
\usepackage{booktabs}

\usepackage{microtype}      % microtypography
\usepackage{xcolor}         % colors
\usepackage[table]{xcolor}
\usepackage{wrapfig}
\usepackage{pifont}

\newcommand{\xmark}{\ding{55}} 
\usepackage{graphicx}  
\usepackage{multirow}  
\usepackage{booktabs}  
\usepackage{array}     
\usepackage{algorithm}
\usepackage{algpseudocode}
\usepackage{amsmath, amssymb}
\usepackage{natbib}
\algrenewcommand\alglinenumber[1]{\normalsize #1:}
\definecolor{Cornsilk}{rgb}{1.0, 0.97, 0.86}
\definecolor{lightorange}{rgb}{0.996, 0.855, 0.643}
\definecolor{lightgray}{rgb}{1.0, 0.827, 0.278}
\usepackage[most]{tcolorbox}
\definecolor{membg}{HTML}{ea6b3c}
\definecolor{memblue2}{HTML}{fcf5ea}
\renewenvironment{abstract}{%
  \begin{tcolorbox}[
    enhanced,
    colback=memblue2,
    colframe=membg,
    boxrule=0.5pt,
    arc=2.5mm,
    left=8mm, right=8mm, top=6mm, bottom=6mm,
    before skip=0pt,
  ]
  \begin{center}
    {\bfseries\color{membg} Abstract}
  \end{center}
  \vspace{3mm}
}{%
  \end{tcolorbox}
}

\hypersetup{colorlinks,linkcolor={blue},citecolor={magenta},urlcolor={blue}}
\newcommand{\github}{\raisebox{-1.5pt}{\includegraphics[height=1.05em]{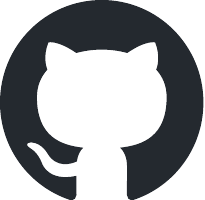}}}
\title{The Mirage of Optimizing Training Policies: Monotonic Inference Policies as the Real Objective for LLM Reinforcement Learning}

\newcommand{\authorshift}{-0.35cm}

\author{
\makebox[\textwidth][c]{\hspace*{\authorshift}%
Jing Liang$^{1,2}$\footnotemark[1] \quad
Hongyao Tang$^{1}$\footnotemark[1] \quad
Yi Ma$^{1}$\footnotemark[1] \quad
Yancheng He$^{2}$\footnotemark[2]\quad
Weixun Wang$^{2}$\footnotemark[2] \quad
\textbf{Xiaoyang Li}$^{2}$%
}\\
\makebox[\textwidth][c]{\hspace*{\authorshift}%
\textbf{Ju Huang}$^{2}$ \quad
\textbf{Wenbo Su}$^{2}$ \quad
\textbf{Jinyi Liu}$^{1}$ \quad
\textbf{Yan Zheng}$^{1}$\footnotemark[2] \quad
\textbf{Jianye Hao}$^{1}$\footnotemark[2] \quad
\textbf{Bo Zheng}$^{2}$%
}\\
\makebox[\textwidth][c]{\hspace*{\authorshift}%
$^{1}$Tianjin University \quad
$^{2}$Alibaba%
}
}

\begin{document}
\renewcommand{\thefootnote}{\fnsymbol{footnote}}
\setcounter{footnote}{0}

\maketitle

\footnotetext[1]{Equal contribution}
\footnotetext[2]{Corresponding authors}

\renewcommand{\thefootnote}{\arabic{footnote}}
\setcounter{footnote}{0}

\begin{abstract}
Reinforcement learning (RL) has gained growing attention in large language model (LLM) post-training, yet RL training remains fragile and can suffer from instability or collapse. 
One vital cause is \textit{training-inference mismatch}: 
LLM adopts separate inference and training engines for generation efficiency and training precision, which in practice exhibits inconsistent probabilities for the same trajectories on training and inference sides, even with synchronized model parameters.
This naturally induces a special type of off-policyness ever existing and poisoning the training.
Prior works have made various efforts in addressing the off-policyness to stabilize the training policies under the mismatch.
In this paper, we point out the \textit{objective misalignment} neglected by existing works that an effective update to the policy in the training engine not necessarily ensures the improvement of the inference policy, i.e., the one used in deployment.
To this end, we propose a new policy optimization objective for LLM RL, named \textbf{Monotonic Inference Policy Improvement (MIPI)}. 
Following this principle, we introduce \textbf{Monotonic Inference Policy Update (MIPU)}, a two-step LLM RL framework that constructs sampler-referenced candidate updates and selectively accepts synchronized candidates using an inference-side gap proxy.
Experiments conducted on two model scales under high mismatch show that \textbf{MIPU} improves average reasoning performance and training stability.

\vspace{0.2cm}

\github{} \textbf{Project Page}: \href{https://anitaleungxx.github.io/MIPU/}{https://anitaleungxx.github.io/MIPU}
\end{abstract}

\section{Introduction}
\label{sec:intro}
With the emergence of reasoning-oriented models such as DeepSeek-R1~\citep{DeepSeek-R1}, Reinforcement Learning (RL) has become an increasingly important paradigm for post-training large language models (LLMs), particularly for improving instruction following, alignment, and reasoning capabilities~\citep{Yu2025DAPO,Open-Reasoner-Zero,qwen3technicalreport}. 
Due to the massive scale of LLM, modern LLM RL pipelines often separate rollout generation from gradient computation: responses are sampled by an inference engine (e.g. vLLM~\citep{vllm} or SGLang~\citep{zheng2024sglang}), whereas log-probabilities and updates are computed by a training engine (e.g. FSDP~\citep{fsdp} or Megatron~\citep{megatron}). 
Even under synchronized parameters, implementation differences in precision, decoding, or serving backends can make the training policy $\textcolor{blue}{\pi}$ and inference policy $\textcolor{red}{\mu}$ assign different probabilities to the same trajectories.
This inevitably creates the notorious \textit{training-inference mismatch} issue~\citep{mismatch1,mismatch2}. 

Existing algorithmic and infrastructure remedies reduce this mismatch by correcting sampler-side ratios~\citep{mismatch1}, filtering unstable samples~\citep{mismatch2}, decaying learning rates~\citep{lrdecay}, or narrowing system-level discrepancies~\citep{li2026qurl,fp16}. 
These works aim to address the mismatch on the training side by improving the stability of the policy optimization process.
However, due to the ever-existing training-inference mismatch, there is no guarantee that effective improvement of the training policy necessarily take the same effect for the inference policy.

\begin{figure}
    \centering
    \includegraphics[width=1\linewidth]{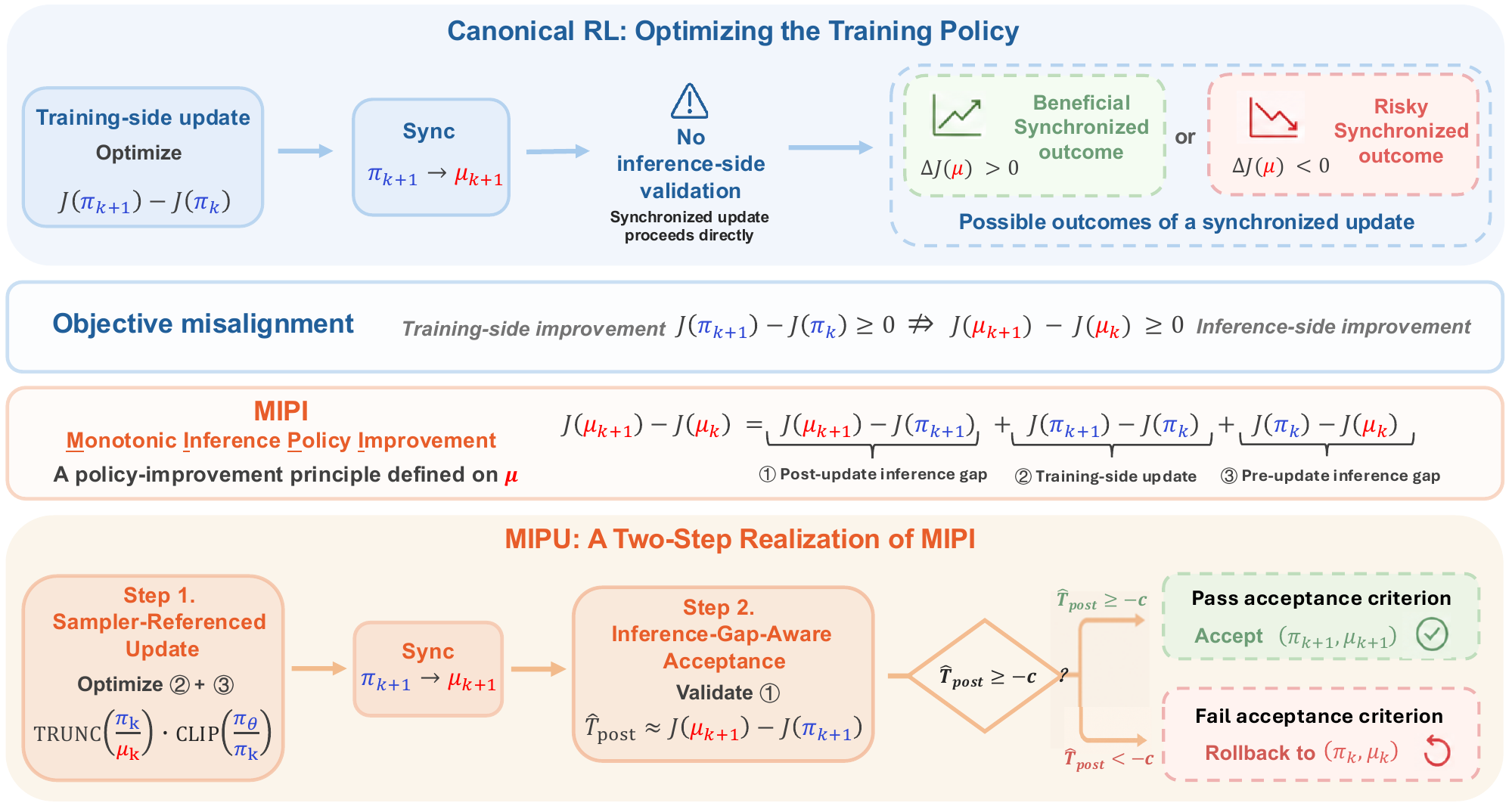}
    \caption{\textbf{Monotonic Inference Policy Update (MIPU) resolves the \textit{Objective Misalignment} issue of LLM RL. }
Canonical LLM RL accepts synchronized updates by a training-side objective, which does not necessarily imply improvement of the inference policy.
Here, $\textcolor{blue}{\pi}$ and $\textcolor{red}{\mu}$ denote the training policy and inference policy respectively, $c$ is a tolerance parameter accounting for proxy noise.
To address this mismatch, we propose a new principle as monotonic improvement on the inference-policy trajectory (the \textbf{MIPI} principle).
\textbf{MIPU} realizes this principle with two steps: \textbf{Step 1 optimizes Terms \textcircled{2}+\textcircled{3}, while Step 2 estimates and validates Term \textcircled{1}, jointly covering all three terms in the MIPI decomposition.}
}
\vspace{-0.5cm}
    \label{fig:method}
\end{figure}

This suggests an \textit{objective-level} view of training-inference mismatch.
Since the policy used for rollout and deployment is induced by the inference engine, a model update should not be judged solely by whether it improves the training policy, i.e., the canonical optimization objective of LLM RL algorithms.
Instead, the more natural question is:
\vspace{-0.1cm}
\begin{center}
\textit{Whether synchronizing this update actually produces a better inference policy ?}
\end{center}
\vspace{-0.1cm}
In face of the objective misalignment revealed above, in this work we aim to break the canonicity of improving the training policy, and propose a new policy optimization principle for LLM RL, named \textbf{\underline{M}onotonic \underline{I}nference \underline{P}olicy \underline{I}mprovement ({MIPI})}: the policy optimization objective should be established to improve the performance of the inference policy monotonically.
Following this principle, we propose a two-step LLM RL framework, called \textbf{\underline{M}onotonic \underline{I}nference \underline{P}olicy \underline{U}pdate (MIPU)}.
As illustrated in~\autoref{fig:method}, the recipe of \textbf{MIPU} is to decompose the \textbf{MIPI} objective (i.e., $J(\textcolor{blue}{\mu_{k+1}})-J(\textcolor{blue}{\mu_k})$) into three terms, and optimize it via a sampler-referenced policy
update step (i.e., the direct `Step 1') and a inference-gap-aware acceptance step (i.e., the indirect `Step 2').
For Step 1, the training engine performs a sampler-referenced policy update to generate a candidate model that theoretically ensures the halfway monotonicity $J(\textcolor{red}{\pi_{k+1}})-J(\textcolor{blue}{\mu_k}) \ge 0$.
For Step 2, the synchronized candidate is evaluated through an inference-gap-aware acceptance criterion, where the accepted update further ensures the remaining half monotonicity $J(\textcolor{blue}{\mu_{k+1}}) -J(\textcolor{red}{\pi_{k+1}}) \ge 0$.

In the experiments, we evaluate \textbf{MIPU} under FP8-quantized rollout, a high-mismatch setting where inference-side quantization amplifies the training-inference discrepancy. 
Across Qwen3-4B and Qwen3-1.7B, \textbf{MIPU} achieves the best average performance and more stable training dynamics. 
Further analysis shows that Step~1 improves candidate updates, while Step~2 filters unreliable synchronized candidates using an inference-side criterion.

The contributions of this paper are summarized as follows:
\begin{itemize}
    \item We propose a new learning objective for LLM RL, i.e., \textbf{MIPI}, which aligns with the purpose de facto and circumvents the training-inference mismatch issue.
    \item We propose a new LLM RL framework, i.e., \textbf{MIPU}, that realizes the \textbf{MIPI} principle in a two-step manner. \textbf{MIPU} allows flexible implementation of both sampler-referenced policy update step and inference-gap-aware update acceptance step.
    \item Experiments under FP8-quantized rollout show that \textbf{MIPU} improves both performance and training stability. And additional analysis verifies the complementary roles of Step~1 and Step~2 in candidate construction and inference-gap-aware acceptance.
\end{itemize}

\section{Preliminaries}
\label{sec:preliminaries}

\paragraph{Reinforcement Learning for LLMs}

Reinforcement Learning (RL)~\citep{Sutton1988ReinforcementLA} is general paradigm for sequential decision-making problems. RL has demonstrated its great ability of optimizing pre-trained LLMs~\citep{trung2024reft,jaech2024openai}.
In this paradigm, sequential token generation is modeled as a Markov Decision Process (MDP) $M = (\mathcal{S}, \mathcal{A}, P, R, \gamma)$,
where a \textbf{state} $s_t = (q, y_{1:t}) \in \mathcal{S}$ is the prompt with the output generated so far, and the \textbf{action} $a_t \in \mathcal{A}$ is the next token selected from the vocabulary $\mathcal{V}$. 
Hence, the \textbf{transition $P(s_{t+1}|s_t, a_t)$} is deterministic in this context.
An episode starts from a prompt $s_0$ (out of a predefined set $\mathcal{D}_0$) and terminates at an end-of-sequence token or by the maximum sequence length $H$.
The \textbf{reward $R(s_t, a_t)$} signal is issued by either a rule-based reward function or a learned reward model.
In the scope of this paper, we consider the verifiable reward function. For any non-terminal timestep $t < T-1$, $R(s_t, a_t)$ is $0$; on completion, the terminal reward, denoted by $R(\tau)$ for the whole sequence, equals $1$ if $\tau$ produces a correct and well-formatted answer and $0$ otherwise.

The \textbf{policy $\pi_\theta(a_t \mid s_t)$} in the MDP is the LLM itself, parameterized by $\theta$, and it defines a probability distribution of next-token generation. 
We use $d^{\pi_{\theta}}_{\tau}$ to denote the distribution of the output sequence $\tau$ generated by $\pi_{\theta}$ and use $d^{\pi_{\theta}}_{s,a}, d^{\pi_{\theta}}_{s}$ for the state-action pairs $(s,a)$ and the state respectively. 
The learning objective of an RL policy is to maximize the reward function, i.e., $\pi^* = \arg \max_{\pi_{\theta}} J(\pi_{\theta})$, where $J(\pi)
    =
    \mathbb{E}_{\tau \sim \pi}
    \left[
        R(\tau)
    \right]$.

\vspace{-0.2cm}

\paragraph{Monotonic Policy Improvement and Proximal Policy Gradient}

Optimizing RL policy according to the vanilla policy gradient $\nabla_{\pi}J(\pi)$ often turns out to be ineffective as the performance is sensitive to the step size. To this end, \citet{SchulmanLAJM15TRPO} revisits the \textit{policy difference}~\citep{KakadeL02CPI} for arbitrary two policies $\pi$ and $\pi^{\prime}$ as defined below:
\begin{equation}
    \Delta(\pi^{\prime},\pi)
    :=
    J(\pi^{\prime}) - J(\pi)
    =
    \mathbb{E}_{s,a \sim d^{\pi^{\prime}}}
    \left[
        A^{\pi}(s,a)
    \right],
    \label{eq:exact_policy_difference}
\end{equation}
where $d^{\pi^{\prime}}$ denotes the discounted state visitation distribution induced by policy $\pi^{\prime}$, and let $A^{\pi}(s,a)$ denote the advantage of taking action $a$ at state $s$ under policy $\pi$. 
Then, a natural idea is to ensure \textit{monotonic policy improvement} during iterative policy optimization.
Formally, given an old policy $\pi_k$, the goal of monotonic policy improvement is to find a new policy $\pi_{k+1}$ such that $J(\pi_{k+1})-J(\pi_k)\geq 0$, i.e., $\Delta(\pi_{k+1}, \pi_k) \ge 0$.

However, directly optimizing~\autoref{eq:exact_policy_difference} is infeasible because it depends on the visitation distribution of the updated policy, i.e., $\pi_{k+1}$. 
\citet{SchulmanLAJM15TRPO} propose Trust-Region Policy Optimization (TRPO) to approximate the exact objective difference by replacing the updated visitation distribution with the old-policy distribution and estimating with the importance sampling ratio:
\begin{equation}
    \widetilde{\Delta}(\pi_{k+1},\pi_k)
    :=
    \mathbb{E}_{s\sim d^{\pi_k},\,a\sim\pi_k(\cdot|s)}
    \left[
        \frac{\pi_{k+1}(a\mid s)}{\pi_k(a\mid s)}
        A^{\pi_k}(s,a)
    \right].
    \label{eq:ratio_surrogate}
\end{equation}
This local surrogate, denoted by $\tilde{\Delta}$, forms the basis of practical proximal policy optimization algorithms. 
TRPO maximizes such a surrogate subject to a KL-based trust-region constraint between the old and updated policies, while Proximal Policy Optimization (PPO)~\citep{SchulmanWDRK17PPO} replaces the constrained optimization with a clipped probability-ratio objective that discourages excessively large policy updates. Based on PPO, Group Relative Policy Optimization (GRPO)~\citep{Shao24GRPO} adopts a group-based advantage estimation which is free of learning the value function.
\vspace{-0.2cm}
\paragraph{Training-Inference Mismatch in LLM RL}
As discussed above, modern LLM RL pipelines may induce different action distributions in the training and inference engines, even when they share synchronized model parameters.
We model this training-inference mismatch by distinguishing the training policy $\textcolor{blue}{\pi}$ from the inference policy $\textcolor{red}{\mu}$: at update step $k$, with the same parameters, $\textcolor{blue}{\pi_k}$ is the policy represented inside the training engine, whereas $\textcolor{red}{\mu_k}$ is the policy actually used for rollout generation and deployment in the inference engine. 
The training-inference mismatch manifests as $\textcolor{blue}{\pi}\neq\textcolor{red}{\mu}$, even though they share the same model parameters.
This mismatch thus results in new learning issues regarding non-stationarity and suboptimality different from the notorious ones in traditional RL~\citep{Hasselt18DRLDeadly,TangB24CHAIN,Wu2026SWD}.
In the following of this paper, we will use the notations to explicitly highlight the distinction between training policies and inference policies.
This distinction provides the basis for our method, where we revisit policy improvement from the perspective of the inference policy rather than the training-side surrogate.

\section{Related Works}
\label{sec:related_works}

\paragraph{Reinforcement learning for LLM post-training}
RL has become a prominent paradigm for LLM post-training, especially for improving reasoning, coding, and long-horizon tasking capabilities.
A major recent direction is reinforcement learning with verifiable rewards (RLVR), where outcome rewards can be obtained from automatically checkable signals such as mathematical answers, code execution, or task success.
Representative works include GRPO~\citep{Shao24GRPO}, DAPO~\citep{Yu2025DAPO}, and GSPO~\citep{gspo}, which develop scalable RL recipes for mathematical and reasoning tasks.
In parallel, recent studies have also explored off-policy or data-reuse variants of reinforced fine-tuning~\citep{remix,zhan2026exgrpo}, highlighting the importance of correcting distributional mismatch during LLM RL.
More recently, RL has also been adopted in agentic model training, where models learn from environment feedback in tool-use, coding, and multi-turn interaction settings.
Examples include GLM-4.5~\citep{glm}, Kimi K2.5~\citep{kimiteam2026kimik25visualagentic}, and ROME~\citep{ROME}.
Together, these works show the broad potential of RL for improving LLM capabilities, while also highlighting the fragility of RL training under long-horizon rollouts, complex optimization dynamics, and system-level implementation details.

\vspace{-0.2cm}
\paragraph{Training-inference mismatch in LLM RL}
Recent work has identified training-inference mismatch as an important source of instability in modern LLM RL systems.
A common line of algorithmic approaches incorporates sampler-side information into the training update.
For example, TIS~\citep{mismatch1} uses the training-to-sampler probability ratio as a clipped correction weight, while MIS~\citep{mismatch2} filters tokens or sequences with extreme mismatch signals.
Learning rate decay~\citep{lrdecay} provides another optimization-side remedy by reducing the update magnitude when mismatch-related instability emerges.
In parallel, infrastructure-level approaches study how system design choices affect training-inference mismatch. 
For example, \citet{fp16} show that numerical precision can be a direct source of mismatch and propose using FP16 rollout to reduce the discrepancy between training and inference engines. 
\citet{li2026qurl} further investigate low-precision RL systems and analyze how quantized rollout affects reasoning-oriented RL training. 
These works reduce mismatch from the system side, while sampler-aware methods such as TIS and MIS correct or filter the training update from the algorithm side. 
Our work is complementary: instead of only reducing mismatch, we study its objective-level consequence and ask whether a synchronized update should be accepted as the next inference policy.

\section{The New Objective: \\
~~~~~~Monotonic Policy Improvement on the Inference Side}
\label{sec:methodology}

As introduced in~\autoref{sec:preliminaries}, training-inference mismatch separates the training policy $\textcolor{blue}{\pi}$ from the inference policy $\textcolor{red}{\mu}$. 
As illustrated in~\autoref{fig:method}, a training-side improvement can still lead to a harmful inference policy after synchronization.
Thus, when $\textcolor{blue}{\pi}\neq\textcolor{red}{\mu}$, improving the training policy does not necessarily imply improving the inference policy:
\begin{equation}
    J(\textcolor{blue}{\pi_{k+1}}) - J(\textcolor{blue}{\pi_k}) \geq 0
    \;\nRightarrow\;
    J(\textcolor{red}{\mu_{k+1}}) - J(\textcolor{red}{\mu_k}) \geq 0.
    \label{eq:mismatch_implication}
\end{equation}

We now instantiate this mismatch in the GRPO objective. 
In GRPO training, the training-side improvement is implemented by the clipped surrogate
\begin{equation}
\begin{aligned}
    \mathcal{J}_{\mathrm{GRPO}}(\theta)
    =
    \mathbb{E}_{x\sim\mathcal{D},\, \{y_i\}_{i=1}^{G} \sim \textcolor{red}{\mu_k}(\cdot\mid x)}
    \left[
        \frac{1}{G}
        \sum_{i=1}^{G}
        \min
        \left(
            r_i(\theta)\hat{A}^{\textcolor{red}{\mu_k}}_i,
            \operatorname{clip}
            \left(
                r_i(\theta),
                1-\epsilon,
                1+\epsilon
            \right)
            \hat{A}^{\textcolor{red}{\mu_k}}_i
        \right)
    \right],
    \label{eq:grpo_objective}
\end{aligned}
\end{equation}
where $r_i(\theta)=\frac{\textcolor{blue}{\pi_\theta}(y_i\mid x)}{\textcolor{blue}{\pi_k}(y_i\mid x)}$ is the training-side probability ratio, and optimizing~\autoref{eq:grpo_objective} yields the next training policy $\textcolor{blue}{\pi_{k+1}}:=\textcolor{blue}{\pi_{\theta^\star}}$. 
The superscript in $\hat{A}^{\textcolor{red}{\mu_k}}_i$ emphasizes that the GRPO advantage is not an explicit estimate of the old training-policy advantage $A^{\textcolor{blue}{\pi_k}}$. 
Instead, it is computed from the response group sampled by the inference policy $\textcolor{red}{\mu_k}$: GRPO normalizes the reward of $y_i$ using the mean and standard deviation of rewards within $\{y_j\}_{j=1}^G\sim\textcolor{red}{\mu_k}(\cdot\mid x)$.

Therefore, beyond the policy-reference mismatch discussed above, GRPO introduces an additional mismatch-induced bias through its advantage estimation.  
Unlike the standard policy surrogate in~\autoref{eq:ratio_surrogate}, whose weighting term is the old-policy advantage $A^{\textcolor{blue}{\pi_k}}$, GRPO weights the update by the group-relative statistic $\hat{A}^{\textcolor{red}{\mu_k}}_i$ induced by inference rollouts. 
When $\textcolor{red}{\mu_k}\neq\textcolor{blue}{\pi_k}$, this statistic is no longer aligned with the advantage of the old training policy. 
Thus, under training-inference mismatch, the standard GRPO update is affected by both a ratio-level mismatch and an advantage-level bias.

More importantly, these mismatches reveal a deeper objective-level misalignment.
The standard GRPO surrogate optimizes a training-policy update, but the resulting synchronized inference policy is the one ultimately used for deployment.
Under training-inference mismatch, the training-policy transition
$\textcolor{blue}{\pi_k}\rightarrow\textcolor{blue}{\pi_{k+1}}$
may differ from the inference-policy transition
$\textcolor{red}{\mu_k}\rightarrow\textcolor{red}{\mu_{k+1}}$.
This motivates us to move beyond correcting individual terms in the training objective and instead ask \textit{what policy improvement should mean when these two transitions differ}.

\subsection{MIPI: Monotonic Inference Policy Improvement}
\label{sec:mipi_objective}

To address this problem, we propose \textbf{Monotonic Inference Policy Improvement (MIPI)}, 
a policy improvement principle for LLM RL under training-inference mismatch.MIPI aims to align the update with the inference-policy transition rather than only with the training-side surrogate.
Formally, MIPI takes the inference-policy improvement
$J(\textcolor{red}{\mu_{k+1}})-J(\textcolor{red}{\mu_k})$
as the target quantity.

To make this principle explicit, we decompose the inference-policy improvement as follows:
\begin{equation}
\begin{aligned}
    J(\textcolor{red}{\mu_{k+1}}) - J(\textcolor{red}{\mu_k})
    =
    \underbrace{
        J(\textcolor{red}{\mu_{k+1}}) - J(\textcolor{blue}{\pi_{k+1}})
    }_{\text{\textcircled{1} post-update inference gap}}
    +
    \underbrace{
        J(\textcolor{blue}{\pi_{k+1}}) - J(\textcolor{blue}{\pi_k})
    }_{\text{\textcircled{2} training-side update}}
    +
    \underbrace{
        J(\textcolor{blue}{\pi_k}) - J(\textcolor{red}{\mu_k})
    }_{\text{\textcircled{3} pre-update inference gap}} .
\end{aligned}
\label{eq:deployed_improvement_decomposition}
\end{equation}

This decomposition provides the formal basis of \textbf{MIPI}.
It separates the inference-policy improvement into a candidate construction part and a post-synchronization verification part.
Accordingly, as illustrated in \autoref{fig:method}, \textbf{MIPU} realizes \textbf{MIPI} through the following two steps:

% \vspace{0.1cm}
\begin{tcolorbox}[
colback=Cornsilk,
colframe=lightgray,
title=\textbf{MIPU: Two-Step Realization of MIPI}
]
\textbf{Step~1: Sampler-referenced policy update.}
\vspace{0.1cm}

    $\triangleright$ The last two terms in~\autoref{eq:deployed_improvement_decomposition} combine into
    $J(\textcolor{blue}{\pi_{k+1}})-J(\textcolor{red}{\mu_k})$.
    We approximate this quantity with a sampler-referenced surrogate to construct a candidate training policy relative to the rollout policy.

\vspace{0.2cm}
\textbf{Step~2: Inference-gap-aware update acceptance.}
\vspace{0.1cm}

    $\triangleright$ The first term
    $J(\textcolor{red}{\mu_{k+1}})-J(\textcolor{blue}{\pi_{k+1}})$
    measures the post-update gap between the candidate training policy and its synchronized inference realization.
    We use this gap as an acceptance signal to decide whether the synchronized candidate should be accepted as the next inference policy. The rejected candidates are rolled back to the previous checkpoint.
\end{tcolorbox}

\subsection{Step 1: Sampler-Referenced Policy Update}
\label{sec:step1_sampler_referenced_update}

Step 1 targets Terms \textcircled{2}+\textcircled{3} in~\autoref{eq:deployed_improvement_decomposition}, which telescope to $J(\textcolor{blue}{\pi_{k+1}})-J(\textcolor{red}{\mu_k})$.
Equivalently, this grouped target can be written as 
$\Delta(\textcolor{blue}{\pi_{k+1}},\textcolor{red}{\mu_k})$.
Following~\autoref{eq:ratio_surrogate}, viewing this target relative to the sampler \(\textcolor{red}{\mu_k}\) leads to the sampler-referenced weighting term \((\textcolor{blue}{\pi_\theta}/\textcolor{red}{\mu_k})A^{\textcolor{red}{\mu_k}}\). As discussed after~\autoref{eq:grpo_objective}, GRPO uses the rollout-induced group-relative statistic \(\hat{A}^{\textcolor{red}{\mu_k}}_i\), so this formulation keeps both the probability ratio and the advantage statistic aligned with the sampler.

By contrast, the standard GRPO surrogate in~\autoref{eq:grpo_objective} clips the training-side ratio 
$\textcolor{blue}{\pi_\theta}/\textcolor{blue}{\pi_k}$, 
although the samples are generated by the inference policy $\textcolor{red}{\mu_k}$. 

Thus, the proximal constraint is centered at the old training policy $\textcolor{blue}{\pi_k}$ rather than at the sampler $\textcolor{red}{\mu_k}$. 
Following the terminology of~\citet{mismatch1}, we refer to this direct sampler-referenced variant as PPO-IS, instead clips the full trainer-to-sampler ratio 
$\rho_i(\theta)=\textcolor{blue}{\pi_\theta}(y_i\mid x)/\textcolor{red}{\mu_k}(y_i\mid x)$. 
However, this full ratio conflates two different effects: the pre-update training-inference mismatch and the current training update:
\begin{equation}
    \rho_i(\theta)
    =
    \underbrace{
        \frac{
            \textcolor{blue}{\pi_k}(y_i\mid x)
        }{
            \textcolor{red}{\mu_k}(y_i\mid x)
        }
    }_{w_i^k:\ \text{pre-update mismatch}}
    \cdot
    \underbrace{
        \frac{
            \textcolor{blue}{\pi_\theta}(y_i\mid x)
        }{
            \textcolor{blue}{\pi_k}(y_i\mid x)
        }
    }_{r_i(\theta):\ \text{current update}} .
    \label{eq:ratio_factorization}
\end{equation}
When $w_i^k$ is already outside the clipping range, PPO-IS can over-constrain the update even when the current update ratio $r_i(\theta)$ remains close to one. 

This factorization suggests separating the pre-update mismatch correction from the clipping of the current update. Following the same terminology, Vanilla-IS follows this idea by using the full correction weight $w_i^k$ while clipping only $r_i(\theta)$, but the unbounded correction can introduce large variance. We therefore adopt the truncated variant TIS, which replaces $w_i^k$ with $\bar{w}*i^k=\min(w_i^k,w*{\max})$ and still applies PPO clipping only to $r_i(\theta)$. Detailed comparisons among PPO-IS, Vanilla-IS, and TIS are provided in~\autoref{app:step1-analysis}.

Accordingly, Step~1 keeps a truncated sampler-referenced correction $\bar{w}_i^k$ while applying PPO-style clipping only to the current-update ratio $r_i(\theta)$. 
The Step-1 surrogate is
\begin{equation}
\begin{aligned}
    \mathcal{J}_{\mathrm{S1}}(\theta)
    =
    \mathbb{E}_{x\sim\mathcal{D},\, \{y_i\}_{i=1}^{G}\sim \textcolor{red}{\mu_k}(\cdot\mid x)}
    \left[
        \frac{1}{G}
        \sum_{i=1}^{G}
        \bar{w}_i^k
        \min
        \left(
            r_i(\theta)\hat{A}^{\textcolor{red}{\mu_k}}_i,
            \operatorname{clip}
            \left(
                r_i(\theta),
                1-\epsilon,
                1+\epsilon
            \right)
            \hat{A}^{\textcolor{red}{\mu_k}}_i
        \right)
    \right].
\end{aligned}
\label{eq:step1_surrogate}
\end{equation}

The truncated weight $\bar{w}_i^k$ retains the sampler-referenced correction while preventing large pre-update discrepancies between $\textcolor{blue}{\pi_k}$ and $\textcolor{red}{\mu_k}$ from dominating the gradient. 
The proximal constraint is still applied only to $r_i(\theta)$, the part of the ratio corresponding to the current training-side update 
$\textcolor{blue}{\pi_k}\rightarrow\textcolor{blue}{\pi_\theta}$. 
Optimizing~\autoref{eq:step1_surrogate} yields a candidate training policy 
$\textcolor{blue}{\pi_{k+1}}:=\textcolor{blue}{\pi_{\theta^\star}}$.

Step 1 provides a sampler-referenced proposal for the combined term 
$J(\textcolor{blue}{\pi_{k+1}})-J(\textcolor{red}{\mu_k})$ 
in~\autoref{eq:deployed_improvement_decomposition}. 
It does not, however, determine whether the synchronized inference policy 
$\textcolor{red}{\mu_{k+1}}$ 
realizes the gain proposed by the trainer. 
This post-update consistency is checked in Step 2.

\subsection{Step 2: Inference-Gap-Aware Update Acceptance}
\label{sec:step2_inference_gap_acceptance}

After Step 1, the candidate training policy 
$\textcolor{blue}{\pi_{k+1}}$ 
is synchronized to the inference engine, producing 
$\textcolor{red}{\mu_{k+1}}$. 
The remaining term in~\autoref{eq:deployed_improvement_decomposition} is the post-update inference gap $T_{\mathrm{post}} = J(\textcolor{red}{\mu_{k+1}}) - J(\textcolor{blue}{\pi_{k+1}})$.

A negative value of $T_{\mathrm{post}}$ means that the inference policy underperforms its training-side counterpart after synchronization. 
In our proxy, this corresponds to an unfavorable mismatch pattern: the inference policy assigns insufficient probability to positive-advantage responses or excessive probability to negative-advantage responses relative to the training policy. 
Such updates are therefore unreliable, even if they appear beneficial on the training side.

By the performance difference identity, the direct form of the post-update gap 
$T_{\mathrm{post}}=\Delta(\textcolor{red}{\mu_{k+1}},\textcolor{blue}{\pi_{k+1}})$ 
requires the advantage under the candidate training policy $\textcolor{blue}{\pi_{k+1}}$, which is not directly available in GRPO-style training. 
We therefore use the reverse identity, which anchors the advantage term at \(\textcolor{red}{\mu_{k+1}}\)\ and makes it compatible with validation rollouts from the synchronized inference policy.
This yields a validation-based proxy with an importance correction from \(\textcolor{red}{\mu_{k+1}}\) to \(\textcolor{blue}{\pi_{k+1}}\):
\begin{equation}
\begin{aligned}
    T_{\mathrm{post}}
    &=
    -
    \Delta(\textcolor{blue}{\pi_{k+1}}, \textcolor{red}{\mu_{k+1}})
    =
    -
    \mathbb{E}_{s,a\sim d^{\textcolor{blue}{\pi_{k+1}}}}
    \left[
        A^{\textcolor{red}{\mu_{k+1}}}(s,a)
    \right] \\
    &\leadsto
    \widehat{T}_{\mathrm{post}}
    =
    -
    \mathbb{E}_{x\sim\mathcal{D}_{\mathrm{val}},\,y_i\sim\textcolor{red}{\mu_{k+1}}}
    \left[
        \rho_i
        \hat{A}^{\textcolor{red}{\mu_{k+1}}}_i
    \right].
\end{aligned}
\label{eq:post_gap_proxy}
\end{equation}
Here, $\hat{A}^{\textcolor{red}{\mu_{k+1}}}_i$ is the group-relative advantage computed from validation responses sampled by $\textcolor{red}{\mu_{k+1}}$. 
The symbol $\leadsto$ emphasizes that $\widehat{T}_{\mathrm{post}}$ is a stable proxy motivated by the reverse identity, rather than an exact estimator of $T_{\mathrm{post}}$. 
Since the reverse form takes expectation under $\textcolor{blue}{\pi_{k+1}}$ while validation responses are sampled from $\textcolor{red}{\mu_{k+1}}$, we introduce an importance weight. 
For each completion $y_i$, we use the length-normalized sequence ratio
\begin{equation}
    \rho_i
    =
    \exp
    \left(
        \frac{1}{T_i}
        \sum_{t=1}^{T_i}
        \log
        \frac{
            \textcolor{blue}{\pi_{k+1}}(y_{i,t}\mid x,y_{i,<t})
        }{
            \textcolor{red}{\mu_{k+1}}(y_{i,t}\mid x,y_{i,<t})
        }
    \right).
    \label{eq:geometric_sequence_ratio}
\end{equation}
This length-normalized ratio serves as a more stabilized importance weight for approximating the reverse-form expectation, avoiding the high variance of full sequence-level importance sampling ratio.

The proxy in~\autoref{eq:post_gap_proxy} serves as a risk signal. 
Since \(\widehat T_{\mathrm{post}}=-\mathbb{E}_{\textcolor{red}{\mu_{k+1}}}[\rho_i\hat{A}^{\textcolor{red}{\mu_{k+1}}}_i]\), a more negative value indicates that \(\rho_i\) is positively associated with the validation advantage statistic. 
In other words, compared with the synchronized inference policy, the candidate training policy assigns relatively higher probability to positive-advantage responses or lower probability to negative-advantage responses.
This suggests that the training-side proposal may not be realized by the inference engine.
It does not prove that the inference policy is worse than the previous one, but marks the update as unreliable under training-inference mismatch.

Therefore, we use the post-update gap proxy as an acceptance test: $\widehat{T}_{\mathrm{post}} \geq -c$,
where $c\geq 0$ is a tolerance parameter for noise in the proxy. 
This rule tolerates mild negative proxy values but rejects candidates with a clear negative post-update gap signal. 
If passing the test, we accept the candidate and set 
$\textcolor{blue}{\pi_k}\leftarrow\textcolor{blue}{\pi_{k+1}}$ and 
$\textcolor{red}{\mu_k}\leftarrow\textcolor{red}{\mu_{k+1}}$. 
Otherwise, we reject the candidate and roll back both the trainer state and the inference engine to the previous checkpoint. The implementation details are provided in~\autoref{app:implementation-details}.

Overall, Step~1 turns the trainer update into a sampler-referenced candidate, and Step~2 filters candidates with a large post-update inference gap.
Algorithm~\autoref{alg:monotonic_inference_update} summarizes the full \textbf{MIPU} procedure, including checkpointing, synchronization, validation, and rollback.
This mechanism does not provide a formal monotonic-improvement guarantee, but it reduces the risk of accumulating updates whose gains are not realized by the inference policy.

\renewcommand{\algorithmiccomment}[1]{$\triangleright$ #1}

\begin{algorithm}[t]
\caption{Monotonic Inference Policy Update}
\label{alg:monotonic_inference_update}
\begin{algorithmic}[1]
\State \textbf{[Input]:} Base model $\pi_{\mathrm{base}}$, group size $G$, clipping range $\epsilon$, tolerance $c$

\State \textbf{[Init]:} Set trainer policy $\textcolor{blue}{\pi_{\theta_0}}=\pi_{\mathrm{base}}$, optimizer state $\Omega_0$, and inference policy $\textcolor{red}{\mu_0}=\mathrm{Sync}(\textcolor{blue}{\pi_{\theta_0}})$
\For{$k=0,1,2,\ldots$}
    \State Save checkpoint $(\theta_k,\Omega_k,\textcolor{red}{\mu_k})$
    \Statex \hspace{0.5cm} \textcolor{orange}{\Comment{Step 1: sampler-referenced proposal}}
    \State Collect a training batch $\mathcal{B}\sim\mathcal{D}$ and collects $G$ samples per prompt from $\textcolor{red}{\mu_k}$
    \State Compute rewards, group-relative advantages $\hat{A}^{\textcolor{red}{\mu_k}}$, and mismatch weights $w^k=\textcolor{blue}{\pi_k}/\textcolor{red}{\mu_k}$ on the training rollouts
    \State Update trainer parameters and optimizer state by maximizing the Step-1 surrogate in~\autoref{eq:step1_surrogate}
    \State Set $\textcolor{blue}{\pi_{k+1}}\leftarrow\textcolor{blue}{\pi_{\theta^\star}}$ and $\textcolor{red}{\mu_{k+1}}\leftarrow\mathrm{Sync}(\textcolor{blue}{\pi_{k+1}})$
    
    \Statex  \hspace{0.5cm} \textcolor{orange}{\Comment{Step 2: monotonic inference acceptance}}
    \State Sample a validation batch $\mathcal{B}_{\mathrm{val}}\sim\mathcal{D}_{\mathrm{val}}$ and roll out $G$ responses per prompt from $\textcolor{red}{\mu_{k+1}}$
\State Compute rewards, advantages $\hat{A}^{\textcolor{red}{\mu_{k+1}}}$, and ratios $\rho_i$ on the validation rollouts
    \State Estimate the post-update inference gap $\widehat{T}_\mathrm{post}$
    \If{$\widehat{T}_\mathrm{post} < -c$}
        \State Reject the update: restore $\theta_{k+1}\leftarrow\theta_k$, $\Omega_{k+1}\leftarrow\Omega_k$, and $\textcolor{red}{\mu_{k+1}}\leftarrow\textcolor{red}{\mu_k}$
    \EndIf
\EndFor
\end{algorithmic}
\end{algorithm}

\section{Experiments}
\label{sec:experiments}

We evaluate \textbf{MIPU} under FP8-quantized rollout, where quantized inference amplifies training-inference mismatch.
Our experiments are organized around three research questions:

\begin{itemize}
    \item \textbf{RQ1:} Does \textbf{MIPU} improve performance and training stability under high training-inference mismatch? 
    (\autoref{subsec:main-res})
    \item \textbf{RQ2:} Do Step~1 and Step~2 play complementary roles in \textbf{MIPU}? 
    (\autoref{subsec:ablation})
    \item \textbf{RQ3:} Is Step~2 effective because of its inference-gap signal, or simply because it rejects more updates? 
    (\autoref{subsec:step2})
\end{itemize}

\subsection{Experiments Setup}

\paragraph{Training.}
We train Qwen3-1.7B and Qwen3-4B under the \emph{FP8-quantized rollout}, where rollout generation uses FP8-quantized inference and therefore induces a high training-inference mismatch.
All analyses beyond the main results are conducted on Qwen3-4B.
The detailed hyperparameter choices are provided in~\autoref{app:hyperparams}.
For training data, we use 
DAPO-Math-17~\citep{Yu2025DAPO} and 
DeepMath-103K~\citep{deepmath} as our training corpora.
Qwen3-1.7B is trained on \texttt{5759} examples filtered from DAPO, while Qwen3-4B is trained on \texttt{1491} examples filtered from DeepMath.
The filtering retains problems that the base model can solve with non-trivial but non-saturated success rates, ensuring that the training set provides meaningful reward variation for RL.

\paragraph{Evaluation.}
We evaluate all methods on five mathematical reasoning benchmarks: MATH-500~\citep{hendrycksmath2021}, AIME24~\citep{AIME24}, AMC23~\citep{AMC23}, Minerva~\citep{minerva}, and OlympiadBench~\citep{olympiadbench}.
We report pass@1 accuracy for all benchmarks.
For the small benchmarks AIME24 and AMC23, we use avg@16 to reduce evaluation variance and improve evaluation reliability.
All reported scores are computed using the same evaluation protocol across methods. 
For diagnostics, we report the inference-training K3-KL (denoted as \textit{Mismatch-K3} in the following figures), computed on rollout tokens sampled from $\mu$. 
Specifically, with $d_t=\log \textcolor{blue}{\pi}(y_t\mid s_t)-\log \textcolor{red}{\mu}(y_t\mid s_t)$, we average $\exp(d_t)-d_t-1$, which estimates $D_{\mathrm{KL}}(\textcolor{red}{\mu}\|\textcolor{blue}{\pi})$.

\paragraph{Baselines.}
Our work focuses on stabilizing the RL objective under training-inference mismatch, which is orthogonal to system-level infrastructure optimization.
We therefore do not compare with infrastructure methods.
For the main comparison, we compare against the standard GRPO baseline and two optimization-side stabilization baselines, MIS and LR-decay.
Since \textbf{MIPU} instantiates Step~1 with a TIS-style correction, we analyze TIS separately as the Step~1-only variant in Sec.~\ref{subsec:ablation}.
The implementation details are in~\autoref{app:details}.

\subsection{Main Results}
\label{subsec:main-res}

\begin{table}[tbhp]
\centering
\caption{\textbf{Pass@1 accuracy (\%)  under FP8-quantized rollout across five mathematical benchmarks.}
\textbf{Bolded} values denote the highest scores in each dataset (i.e., column).
The \textit{\textbf{Stable}} column indicates whether the method avoids collapse or sharp degradation during continued training after reaching high performance.
On both Qwen3-4B and Qwen3-1.7B, \textbf{MIPU} achieves better average scores than other methods, while exhibiting more stable training dynamics.}
\vspace{0.2cm}
\label{tab:results-fp8}
\resizebox{\textwidth}{!}{ 
\begin{tabular}{l l ccccc c | c}
\toprule
\textbf{Model} & \textbf{Method} & \textbf{MATH} & \textbf{AIME} & \textbf{Olympiad} & \textbf{Minerva} & \textbf{AMC23} & \textbf{Avg.} & \textbf{\textit{Stable}} \\
\midrule
\multirow{4}{*}{\textbf{Qwen3-4B}} 
& Baseline  & 89.34 & 42.00 & 64.89 & 43.39 & 82.50 & 64.42 & \xmark \\
& MIS       & 90.95 & 38.44 & 62.50 & 44.12 & 81.09 & 63.42 & \xmark \\
& LR-decay  & 90.34 & \textbf{44.00} & 67.26 & 43.75 & 82.97 & 65.66 & \xmark \\
& \cellcolor{orange!20}Ours & \cellcolor{orange!20}\textbf{91.15} & \cellcolor{orange!20}43.56 & \cellcolor{orange!20}\textbf{67.86} & \cellcolor{orange!20}\textbf{45.96} & \cellcolor{orange!20}\textbf{85.00} & \cellcolor{orange!20}\textbf{66.71} & \cellcolor{orange!20}\checkmark \\
\midrule
\multirow{4}{*}{\textbf{Qwen3-1.7B}} 
& Baseline  & 83.10 & \textbf{25.33} & 56.55 & 31.68 & 57.66 & 50.86 & \xmark \\
& MIS       & 81.29 & 24.67 & 58.33 & \textbf{34.19} & 60.16 & 51.73 & \xmark \\
& LR-decay  & 82.09 & 26.00 & 58.93 & 28.68 & \textbf{65.47} & 52.23 & \xmark \\
& \cellcolor{orange!20}Ours & \cellcolor{orange!20}\textbf{86.52} & \cellcolor{orange!20}24.67 & \cellcolor{orange!20}\textbf{59.52} & \cellcolor{orange!20}33.82 & \cellcolor{orange!20}65.31 & \cellcolor{orange!20}\textbf{53.97} & \cellcolor{orange!20}\checkmark \\
\bottomrule
\end{tabular}
}
\end{table}

\begin{wrapfigure}{r}{0.5\linewidth}
    \vspace{-0.6cm}
    \centering
    \includegraphics[width=\linewidth]{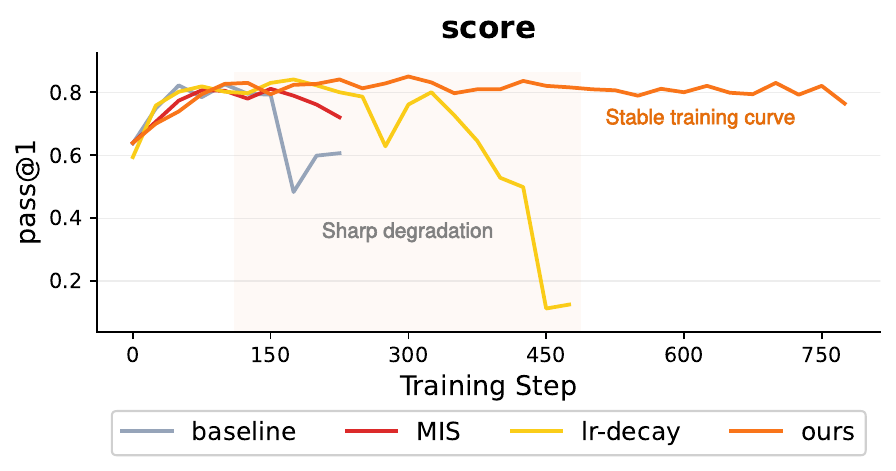}
    \caption{\textbf{Performance of different methods under FP8-quantized rollout. }Compared methods show unstable training dynamics and may suffer from sharp performance drops, while MIPU maintains a stable score trajectory.}
    \vspace{-0.4cm}
    \label{fig:main}
\end{wrapfigure}

\autoref{tab:results-fp8} reports the main results under FP8-quantized rollout, the high-mismatch setting where quantized inference amplifies the gap between the training policy $\pi$ and the inference policy $\mu$.
On both model scales, \textbf{MIPU} achieves the best average performance, reaching \textbf{$66.71\%$} on Qwen3-4B and \textbf{$53.97\%$} on Qwen3-1.7B.
For Qwen3-4B, the gains are most visible on AMC23 ($85.00\%$) and Minerva ($45.96\%$); for Qwen3-1.7B, the gains are most visible on MATH-500 ($86.52\%$) and OlympiadBench ($59.52\%$).
Beyond peak accuracy, the Stable column in~\autoref{tab:results-fp8} shows that \textbf{MIPU} remains stable until the end of training, while several baselines suffer from collapse or sharp degradation.

It is worth noting that some baselines can achieve competitive intermediate performance, but their gains are not sustained under continued training.
This distinction is important in high-mismatch RL, where a transient peak may be followed by rapid degradation once mismatch-driven errors accumulate.
Therefore,~\autoref{fig:main} further visualizes the training dynamics under the Qwen3-4B FP8-quantized rollout.
Baseline, MIS, and LR-decay can reach reasonable intermediate performance but later degrade during continued training.
In contrast, \textbf{MIPU} maintains a stable trajectory after reaching high performance.
Together, the table and training curve show that \textbf{MIPU} improves both peak performance and long-horizon training stability in the high-mismatch FP8-quantized rollout.

\subsection{Ablation Studies}
\label{subsec:ablation}

We ablate \textbf{MIPU} under the representative Qwen3-4B FP8-quantized rollout. We compare the baseline GRPO update, Step~1 only, Step~2 only, and the full Step~1+Step~2 framework. This ablation is designed to disentangle two complementary roles: Step~1 constructs sampler-referenced candidate updates, while Step~2 decides whether the synchronized inference policy should accept the candidate according to $\widehat{T}_{\mathrm{post}}$.

\begin{table}[tbhp]
\centering
\caption{\textbf{Ablation study under FP8-quantized rollout of Qwen3-4B.} 
We report pass@1 accuracy (\%) on each benchmark.
The full method achieves the best overall performance by combining sampler-referenced candidate construction (Step~1) with inference-gap-aware candidate acceptance (Step~2), showing that the two components are complementary.
\textbf{Bolded} values denote the highest scores in each dataset (i.e., column).}
\vspace{0.2cm}
\label{tab:ablation}
\begin{tabular}{lcccccc}
\toprule
\textbf{Method} & \textbf{MATH 500} & \textbf{AIME 24} & \textbf{Olympiad} & \textbf{Minerva} & \textbf{AMC23} & \textbf{Avg.} \\
\midrule
Baseline        & 89.34 & 42.00 & 64.89 & 43.39 & 82.50 & 64.42 \\
+ Step 1    & 90.34 & 41.11 & \textbf{68.45} & 44.85 & 82.03 & 65.36 \\
+ Step 2    & 90.34 & 40.44 & 64.88 & 43.38 & 75.00 & 62.81 \\
\rowcolor{orange!20}Ours & \textbf{91.15} & \textbf{43.56} & 67.86 & \textbf{45.96} & \textbf{85.00} & \textbf{66.71} \\
\bottomrule
\end{tabular}
\end{table}

\begin{figure}[tbhp]
    \centering
    \includegraphics[width=\linewidth]{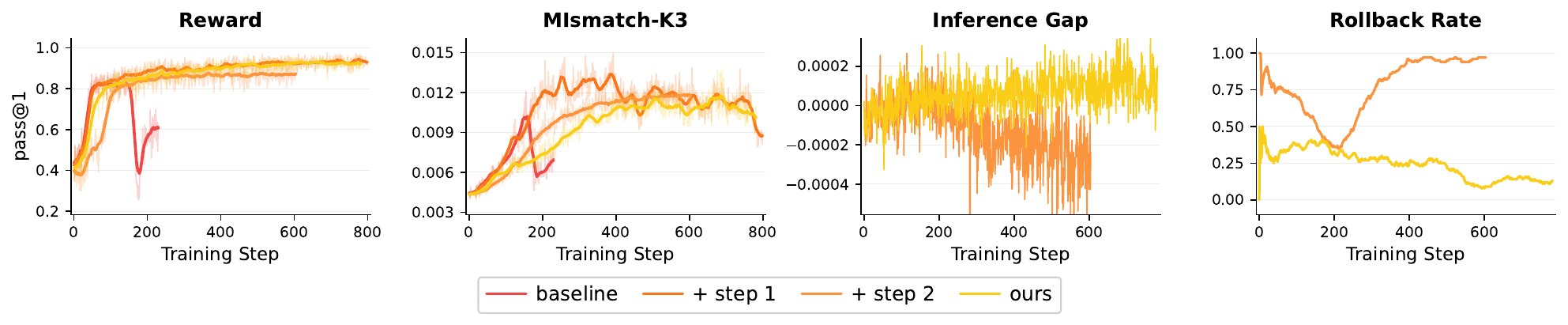}
    \caption{\textbf{Training curves for ablation studies under FP8-quantized rollout.}
We show the training score, the inference-training K3-KL, $\widehat{T}_{\mathrm{post}}$ (i.e., inference gap) and the rollback rate computed over a 100-step moving window.
Step~1 improves the candidate update direction, while Step~2 introduces inference-gap-aware acceptance to filter unreliable synchronized candidates.
The full method obtains stronger performance with a more controlled inference-policy trajectory.
}
\label{fig:ablation}
\end{figure}

\autoref{tab:ablation} and~\autoref{fig:ablation} show that Step~1 and Step~2 address different failure modes.
The baseline update is centered at the training policy $\pi_k$ even though rollouts are sampled from $\mu_k$, which makes the synchronized inference policy vulnerable to FP8-induced mismatch and can eventually lead to collapse.
Step~2 alone can prevent collapse by rejecting candidates with large negative $\widehat{T}_{\mathrm{post}}$, but it cannot improve the quality of the underlying baseline update.
Thus, when the baseline rarely proposes useful candidates, Step~2 mostly keeps the previous inference policy rather than creating further improvement.
Its acceptance rule improves safety, but it cannot compensate for poor candidate quality when the proposal update itself remains biased by the training-inference mismatch.

Step~1 improves learning by constructing sampler-referenced candidate updates before synchronization.
However, Step~1 alone accepts every synchronized candidate as the next inference policy, so mismatch fluctuations can still accumulate.
The full method combines the two roles: Step~1 produces better candidate updates, and Step~2 filters candidates whose post-update inference gap suggests that the gain may not be realized by $\mu_{k+1}$.

In summary, Step~1 and Step~2 address different bottlenecks: Step~1 improves the candidate update direction, while Step~2 controls whether the synchronized candidate should be accepted as the next inference policy. Strong performance requires corrected updates from Step~1 together with inference-gap-aware acceptance from Step~2.
This division of roles explains why the full method improves both the final score and the stability of the training trajectory, rather than merely shifting the best checkpoint along an unstable run.

\subsection{Analysis of Step 2: Inference-Gap-Aware Acceptance}
\label{subsec:step2}

\begin{figure}[tbhp]
    \centering
    \begin{subfigure}[t]{0.6\linewidth}
        \centering
        \includegraphics[width=\linewidth]{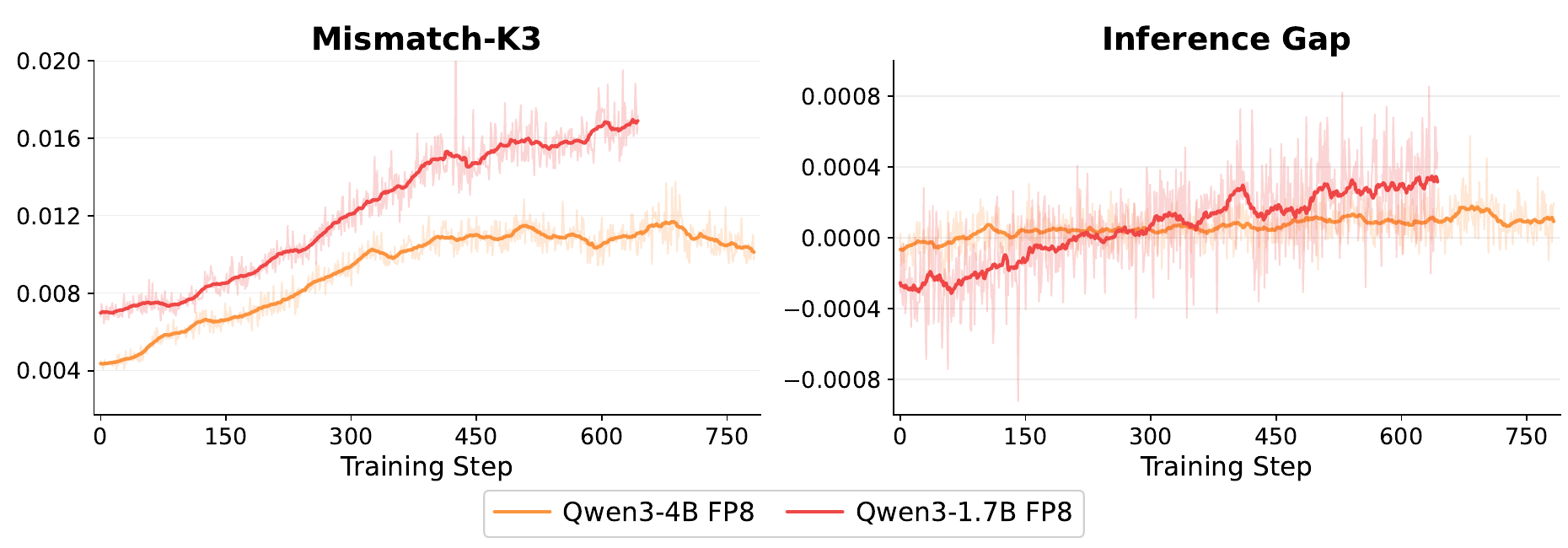}
        \caption{Inference-training K3-KL and inference gap.}
        \label{fig:term1-mismatch}
    \end{subfigure}
    \hfill
    \begin{subfigure}[t]{0.32\linewidth}
        \centering
        \includegraphics[width=\linewidth]{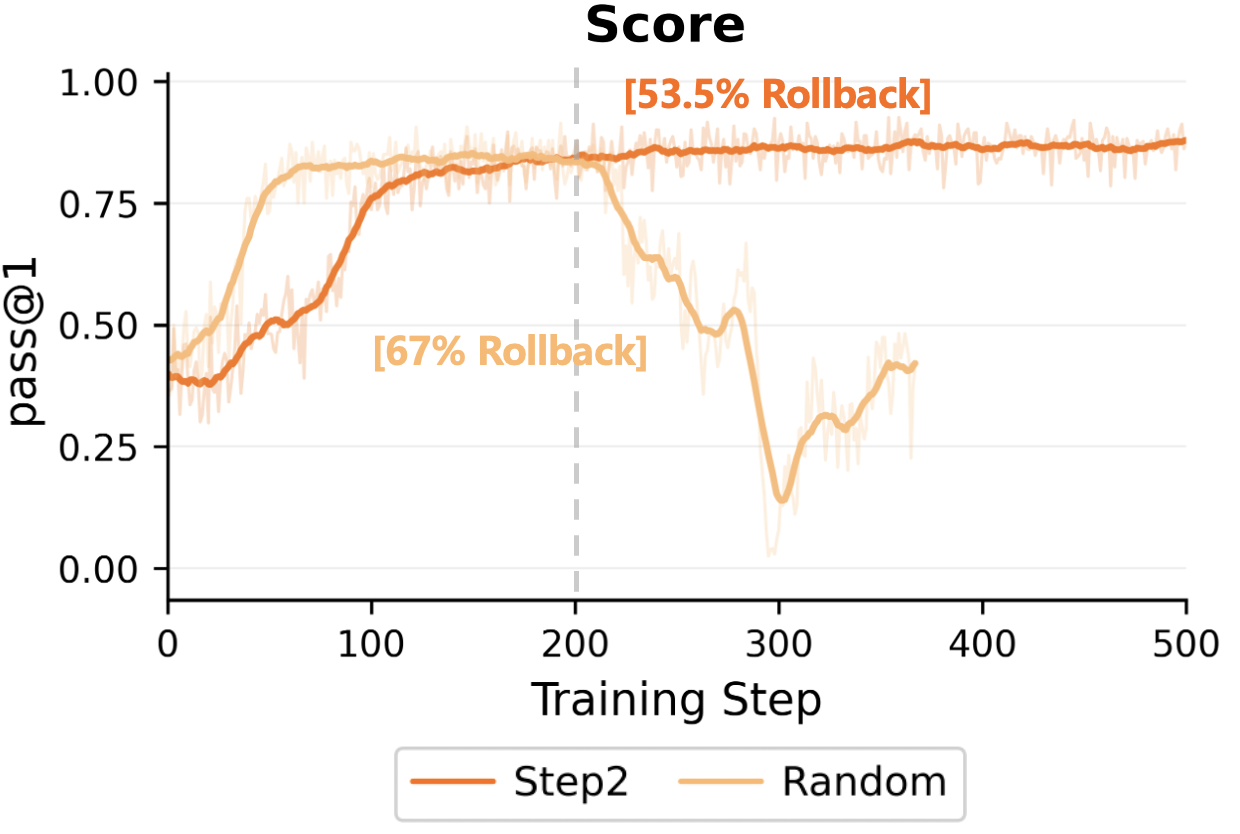}
        \caption{Step~2 vs. random rollback.}
        \label{fig:random-rollback}
    \end{subfigure}
    \caption{
    \textbf{(a) Inference-training K3-KL and $\widehat{T}_{\mathrm{post}}$ (i.e., inference gap) under FP8-quantized rollout.} 
    Qwen3-1.7B exhibits larger mismatch and a more volatile $\widehat{T}_{\mathrm{post}}$ than Qwen3-4B.
    \textbf{(b) Comparison between inference-gap-aware Step~2 acceptance and a random rollback control.}
    Random rollback rejects more updates, applying fewer effective policy changes, but still collapses.
    }
    \label{fig:step2-analysis}
\end{figure}

\paragraph{Post-update gap as an inference-side signal.}
We first inspect whether the post-update gap proxy $\widehat{T}_{\mathrm{post}}$ carries a meaningful mismatch signal.
As shown in~\autoref{fig:term1-mismatch}, under FP8-quantized rollout, the smaller Qwen3-1.7B model exhibits a larger training-inference mismatch than Qwen3-4B.
Consistently, its $\widehat{T}_{\mathrm{post}}$ has a larger magnitude and oscillates more strongly, indicating a more volatile inference-side gap under quantized rollout.
Moreover, both models show a training-dependent pattern: the mismatch increases during training, and $\widehat{T}_{\mathrm{post}}$ also shifts from initially low values toward higher values.
These observations suggest the use of Step~2 in high-mismatch settings, where $\widehat{T}_{\mathrm{post}}$ carries a structured inference-side signal rather than behaving as uninformative boundary-level noise.
This also motivates using \(\widehat{T}_{\mathrm{post}}\)\ as an acceptance signal: it reflects whether the synchronized inference policy remains consistent with the candidate update in a mismatch-relevant direction, rather than merely measuring the magnitude of the mismatch.

\paragraph{Signal-conditioned rollback.}
We next test whether the benefit of Step~2 can be explained merely by reducing the number of accepted updates.
We compare it with a random rollback control that rejects candidate updates without using $\widehat{T}_{\mathrm{post}}$.
Since the Step~2-only run, without Step~1 correction, has an overall rollback rate of around $70\%$ within the first $500$ steps, we set the random rollback probability to $70\%$.
In the comparison window shown in~\autoref{fig:random-rollback}, random rollback is even more conservative than Step~2, rejecting $67.0\%$ of candidate updates compared with $53.5\%$ for Step~2.

Despite rejecting more updates, random rollback still collapses after reaching a transient peak, whereas Step~2 maintains a stable score trajectory.
This shows that Step~2 is not a generic update-sparsification mechanism.
Its stability comes from conditioning the acceptance decision on the post-update gap proxy $\widehat{T}_{\mathrm{post}}$, which helps filter unreliable synchronized candidates rather than randomly reducing the number of accepted updates.
Random rollback is more conservative in terms of accepted-update count, but it cannot distinguish harmful synchronized candidates from benign ones.
By contrast, Step~2 uses $\widehat{T}_{\mathrm{post}}$ to identify candidates whose training-side gains are unlikely to be realized by the inference policy. 
Overly conservative rejection may preserve a stale inference policy, whereas inference-gap-aware acceptance allows useful candidates to pass while filtering candidates with mismatch-relevant risk.

We further analyze the sensitivity of the acceptance tolerance in~\autoref{app:tolerance}, showing that Step~2 requires calibrated acceptance rather than simply stricter rejection.

\section{Conclusion}
\label{sec:conclusion}
We revisit training-inference mismatch in LLM reinforcement learning from an objective-level perspective. 
Rather than treating mismatch only as a system discrepancy to be reduced, we show that it changes the object of policy improvement: standard RL updates optimize the training policy, while deployment depends on the inference policy. 
Based on this view, we formulate Monotonic Inference Policy Improvement (\textbf{MIPI}) and propose Monotonic Inference Policy Update (\textbf{MIPU}), which realizes \textbf{MIPI} through sampler-referenced candidate construction and inference-gap-aware candidate acceptance. 
Experiments under FP8-quantized rollout show that \textbf{MIPU} improves average reasoning performance and training stability, and further analysis confirms the rationality of Step~2.
These results suggest that training-inference mismatch should not be treated only as a low-level system discrepancy, but also as a change in the objective of policy improvement.

\paragraph{Limitations.}
Due to significant computational constraints, our experiments are currently limited to moderate-scale models. 
Future work should examine whether the same mismatch patterns also appear in larger models and more diverse RL training systems.
In addition, Step~2 should be viewed as a flexible design space rather than a fixed implementation. 
Our validation-based post-update gap proxy provides one effective instantiation for evaluating synchronized candidates from the deployment side. 
Future work may further improve this component with more efficient estimators, stronger acceptance criteria, or optimization schemes that directly leverage the post-update inference gap.

\bibliography{ref}
\bibliographystyle{arxiv}

%%%%%%%%%%%%%%%%%%%%%%%%%%%%%%%%%%%%%%%%%%%%%%%%%%%%%%%%%%%%
\newpage
\appendix

\section{Implement Details}
\label{app:details}

\subsection{Hyperparmeters}
\label{app:hyperparams}
The major hyperparameter choices are shown in~\autoref{tab:hypers}.

\begin{table}[h]
\caption{\textbf{Hyperparameter setups.}}
\vspace{0.25cm}
\centering
\begin{tabular}{l|l}
\toprule
\textbf{Parameter} & \textbf{Value} \\
\midrule
\multicolumn{2}{l}{\textit{Decoding Configuration}} \\
\midrule
training temperature & 1.0\\
evaluation temperature & 0.7\\
top-p & 1.0\\
top-k & -1\\
response\_length & 8192 \\
prompt\_length & 512 \\
\midrule
\multicolumn{2}{l}{\textit{Training Configuration}} \\
\midrule
learning\_rate & 1e-6 \\
num\_return\_sequences\_in\_group & 8\\
train\_batch\_size & 64 \\
dual\_clip\_loss & true \\
kl\_loss\_coef & 0.001 \\
gradient\_accumulation\_steps & 32 \\
\midrule
\multicolumn{2}{l}{\textit{Infrastructure Configuration}} \\
\midrule
GPUs & H100 * 8\\
RL framwork & ROLL~\citep{roll} \\
Training engine & Megatron \\
Inference engine & vLLM \\
\bottomrule
\end{tabular}
\label{tab:hypers}

\end{table}

\subsection{Implementation Details}
\label{app:implementation-details}
\paragraph{Baseline.}
We use vanilla GRPO with the same rollout, reward, and optimization settings as other methods. 
In implementation, we additionally use the dual-clipped policy loss. 
For a token-level ratio $r_{i,t}(\theta)=\pi_\theta(y_{i,t}\mid x,y_{i,<t})/\pi_k(y_{i,t}\mid x,y_{i,<t})$ and group-relative advantage $\hat{A}_i$, the standard clipped objective is
$\ell^{\mathrm{clip}}_{i,t}=\min\!\left(r_{i,t}(\theta)\hat{A}_i,\operatorname{clip}(r_{i,t}(\theta),1-\epsilon,1+\epsilon)\hat{A}_i\right)$.
The dual-clipped objective is then
\[
\ell^{\mathrm{dc}}_{i,t}
=
\begin{cases}
\ell^{\mathrm{clip}}_{i,t}, & \hat{A}_i \ge 0,\\
\max\!\left(\ell^{\mathrm{clip}}_{i,t}, (1+2\epsilon)\hat{A}_i\right), & \hat{A}_i < 0,
\end{cases}
\]
where we set $\epsilon=0.2$. 
The baseline maximizes the average dual-clipped GRPO objective over sampled responses.

\paragraph{Ours.}
For Step~1, we instantiate the sampler-referenced update with token-level truncated importance weights. 
Specifically, for each generated token, we compute the training-to-inference mismatch weight $w^k_{i,t}$ and set $w_{\max}=2$. 
The weight is applied after the dual-clipped GRPO objective is computed, yielding the Step~1 token-level objective
\[
\mathcal{J}_{\mathrm{S1}}
=
\mathbb{E}
\left[
\frac{1}{G}
\sum_{i=1}^{G}
\frac{1}{T_i}
\sum_{t=1}^{T_i}
\bar{w}^k_{i,t}\,
\ell^{\mathrm{dc}}_{i,t}
\right].
\]
Vanilla-IS follows the same implementation except that it uses the untruncated weight $w^k_{i,t}$ instead of the truncted $\bar{w}^k_{i,t}$.

For Step~2, after the candidate training policy $\pi_{k+1}$ is synchronized to the inference engine as $\mu_{k+1}$, we estimate the post-update inference gap proxy $\widehat{T}_{\mathrm{post}}$ on a validation batch. 
The candidate is accepted if $\widehat{T}_{\mathrm{post}}\ge -c_t$, where $c_t\ge 0$ is a dynamic acceptance tolerance parameter. 
Motivated by the observation in~\autoref{fig:term1-mismatch} that $\widehat{T}_{\mathrm{post}}$ is initially low and then gradually increases under FP8 rollout, we use a larger tolerance in the early stage to avoid over-rejecting useful candidates. 
The threshold is linearly annealed during the first $100$ steps:
\[
c_t
=
c_{\mathrm{start}}
+
\frac{\min(t,100)}{100}
\left(
c_{\mathrm{end}}-c_{\mathrm{start}}
\right).
\]
For Qwen3-1.7B, we set $c_{\mathrm{start}}=4\times 10^{-3}$ and $c_{\mathrm{end}}=1\times 10^{-3}$.
For Qwen3-4B, we set $c_{\mathrm{start}}=1\times 10^{-3}$ and $c_{\mathrm{end}}=0$.
After the first $100$ steps, the tolerance is fixed at $\tau_{\mathrm{end}}$.
We further analyze the effect of different tolerance choices in~\autoref{app:tolerance}.

\paragraph{MIS.}
We implement MIS as a token-level mismatch-filtering baseline. 
For each token, we compute the same training-to-inference ratio $w^k_{i,t}$. 
Tokens whose mismatch weight exceeds the upper bound are masked out:
\[
m^k_{i,t}
=
\mathbbm{1}
\left[
w^k_{i,t}\le \bar{w}
\right],
\qquad
\bar{w}=2.
\]
The MIS objective applies this mask to the dual-clipped GRPO objective:
\[
\mathcal{J}_{\mathrm{MIS}}
=
\mathbb{E}
\left[
\frac{1}{G}
\sum_{i=1}^{G}
\frac{1}{T_i}
\sum_{t=1}^{T_i}
m^k_{i,t}\,
\ell^{\mathrm{dc}}_{i,t}
\right].
\]

\paragraph{LR-decay.}
We implement LR-decay following the original decay rule. 
Let $\eta_t$ denote the learning rate at training step $t$. 
At every decay interval $T_{\mathrm{decay}}$, the learning rate is halved until it reaches a lower bound $\eta_\infty$:
\[
\eta_t
=
\begin{cases}
\max\!\left(\eta_{t-1}/2,\eta_\infty\right),
& t \equiv 0 \pmod{T_{\mathrm{decay}}},\\
\eta_{t-1},
& \text{otherwise}.
\end{cases}
\]
The model parameters are then updated using $\eta_t$. 
Following the authors' guidance, we set $T_{\mathrm{decay}}=249$ for Qwen3-1.7B and $T_{\mathrm{decay}}=234$ for Qwen3-4B under FP8 rollout.

\section{Analysis of Step 1: Sampler-Referenced Update}
\label{app:step1-analysis}

\begin{figure}[tbhp]
    \centering
    \includegraphics[width=\linewidth]{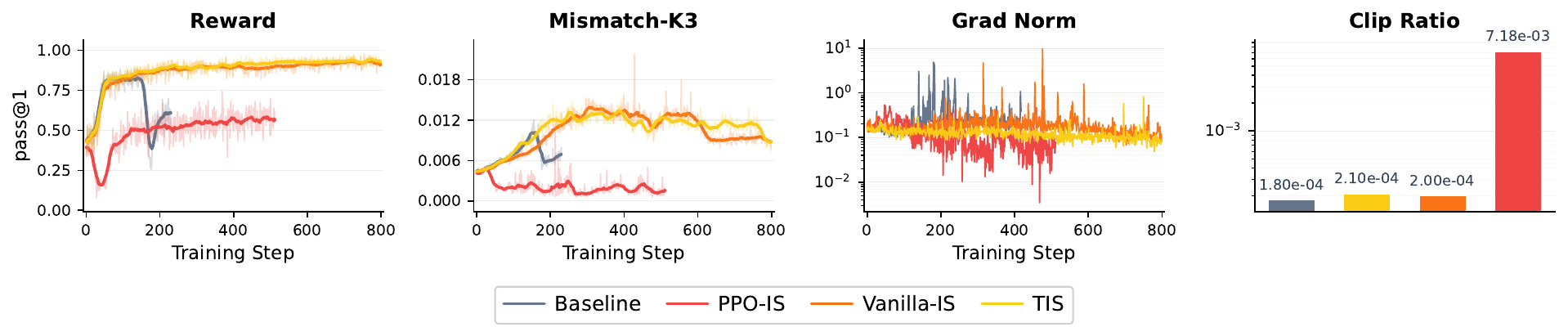}
    \caption{\textbf{Step~1 implementation analysis under the Qwen3-4B FP8-quantized rollout.} Comparison of PPO-IS, Vanilla-IS, and TIS in terms of performance, gradient norm, inference-training K3-KL, and clip ratio.}
    \label{fig:step1}
\end{figure}

Step~1 implements the sampler-referenced update derived in~\autoref{eq:ratio_factorization}. 
For readability, we omit token or trajectory arguments in this section, and all ratios are evaluated on sampled rollouts. 
A direct implementation is PPO-IS, which clips the total trainer-to-sampler ratio:
\[
\textbf{PPO-IS:}
\qquad
\omega_{\mathrm{PPO\text{-}IS}}
=
\operatorname{clip}\!\left(
\frac{\textcolor{blue}{\pi_{\theta}}}{\textcolor{red}{\mu_k}},
1-\epsilon,
1+\epsilon
\right).
\]
Although this ratio directly references the sampler, it mixes the pre-update mismatch between $\textcolor{blue}{\pi_k}$ and $\textcolor{red}{\mu_k}$ with the current update from $\textcolor{blue}{\pi_k}$ to $\textcolor{blue}{\pi_{\theta}}$. 
As a result, the ratio can already deviate substantially from $1$ even before the current update is applied. 
This makes the PPO clipping overly active and therefore less informative, because the trust-region constraint is no longer centered on the current training-side update itself. 
This phenomenon is also consistent with the analysis in \citet{mismatch1}. 
Empirically,~\autoref{fig:step1}(b) shows that PPO-IS has a much larger clipped ratio than the other variants, and~\autoref{fig:step1}(a) shows that, although it reduces the measured inference-training KL, it fails to produce a sustained upward learning trajectory.

We therefore use the factorization
\[
\frac{\textcolor{blue}{\pi_{\theta}}}{\textcolor{red}{\mu_k}}
=
\underbrace{
\frac{\textcolor{blue}{\pi_k}}{\textcolor{red}{\mu_k}}
}_{\text{pre-update mismatch correction}}
\cdot
\underbrace{
\frac{\textcolor{blue}{\pi_{\theta}}}{\textcolor{blue}{\pi_k}}
}_{\text{current update ratio}},
\]
which separates the pre-update mismatch correction from the current training-side update. 
This leads to two decomposed variants:
\begin{gather*}
\textbf{Vanilla-IS:}
\qquad
\omega_{\mathrm{Vanilla\text{-}IS}}
=
\frac{\textcolor{blue}{\pi_k}}{\textcolor{red}{\mu_k}}
\cdot
\operatorname{clip}\!\left(
\frac{\textcolor{blue}{\pi_{\theta}}}{\textcolor{blue}{\pi_k}},
1-\epsilon,
1+\epsilon
\right),
\\[0.6em]
\textbf{TIS:}
\qquad
\omega_{\mathrm{TIS}}
=
\min\!\left(
\frac{\textcolor{blue}{\pi_k}}{\textcolor{red}{\mu_k}},
C
\right)
\cdot
\operatorname{clip}\!\left(
\frac{\textcolor{blue}{\pi_{\theta}}}{\textcolor{blue}{\pi_k}},
1-\epsilon,
1+\epsilon
\right).
\end{gather*}

Vanilla-IS keeps the trust-region constraint on the current update ratio $\textcolor{blue}{\pi_{\theta}}/\textcolor{blue}{\pi_k}$, which is more suitable than directly clipping $\textcolor{blue}{\pi_{\theta}}/\textcolor{red}{\mu_k}$. 
However, its mismatch-correction weight $\textcolor{blue}{\pi_k}/\textcolor{red}{\mu_k}$ is unbounded. 
When a rollout token is sampled with low probability under $\textcolor{red}{\mu_k}$, this weight can become large and significantly amplify gradient variance, as also discussed in \citet{mismatch1}. 
This explains why Vanilla-IS recovers a clearer upward learning trend than PPO-IS, but still exhibits larger gradient-norm spikes in~\autoref{fig:step1}(a).

TIS further truncates the mismatch-correction weight while preserving the decomposed form. 
As shown in~\autoref{fig:step1}, this yields the best stability-performance trade-off among the Step~1 variants: compared with PPO-IS, it avoids overly aggressive clipping, and compared with Vanilla-IS, it controls variance amplification from large mismatch weights. 
We therefore use TIS to construct the Step~1 candidate update in the full method.

In summary, directly clipping $\textcolor{blue}{\pi_{\theta}}/\textcolor{red}{\mu_k}$ makes the trust-region constraint depend on the pre-existing mismatch, which can render the update overly conservative. 
The decomposed form separates mismatch correction from the current update, and the truncated version used by TIS further controls variance, making it the most effective Step~1 implementation in our setting.

\section{Analysis of the Tolerance Paremeter $c$}

\label{app:tolerance}

We further study how the acceptance tolerance affects Step~2. 
In the main method, the synchronized candidate is accepted if 
$\widehat{T}_{\mathrm{post}}\ge -c$, where $c\ge 0$ is a tolerance parameter. 
For sensitivity analysis, we keep the same boundary form $\widehat{T}_{\mathrm{post}}\ge -c$ and also test $c<0$, which corresponds to a stricter-than-zero acceptance rule rather than a tolerance. 
Thus, $c>0$ allows mild negative proxy values, $c=0$ gives strict non-regression under the proxy, and $c<0$ requires a positive margin.

\begin{figure}[tbhp]
    \centering
    \includegraphics[width=\linewidth]{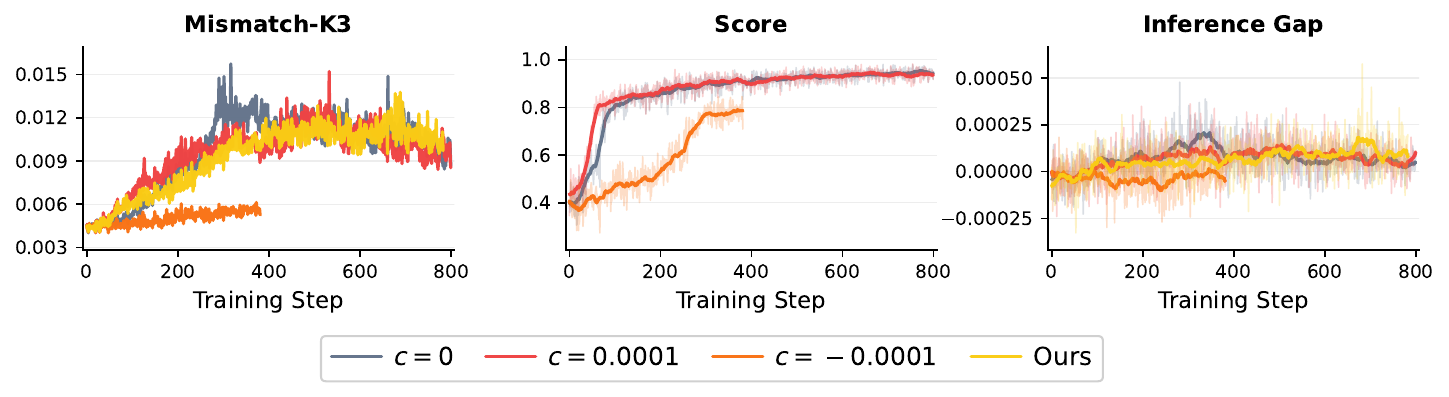}
    \caption{\textbf{Sensitivity to the acceptance tolerance $c$ under the Qwen3-4B FP8-quantized rollout} in terms of inference-training K3-KL, training score, and $\widehat{T}_{\mathrm{post}}$.}
    \label{fig:tolerance}
\end{figure}

\autoref{fig:tolerance} shows that stricter acceptance is not always beneficial. 
When the acceptance rule is too strict, Step~2 rejects many early candidates. 
This is especially harmful because $\widehat{T}_{\mathrm{post}}$ is initially low under FP8-quantized rollout and gradually increases during training, as observed in~\autoref{fig:step2-analysis}(a). 
As a result, overly strict acceptance slows down learning and leads to a lower performance plateau. 
In particular, the stricter setting enters persistent rollback after around $280$ steps, suggesting that conservative rejection alone cannot provide sufficient policy improvement.

These results show that rollback rate itself is not the objective. 
Although rejecting more updates may appear safer, excessive rollback can prevent the inference policy from moving to a better region. 
The dynamic tolerance used in \textbf{MIPU} provides a calibrated acceptance rule: it starts with a larger tolerance to allow useful early candidates, and gradually tightens the criterion as the post-update gap becomes more stable. 
This balances early learning efficiency with later-stage protection against mismatch-driven unreliable updates.

\section{Licenses}
\label{app:licenses}

The licenses of used assets in this paper are listed as follows:
\begin{itemize}
    \item MATH500: MIT License
    \item AIME24: Apache-2.0 License
    \item AMC23: No License
    \item Minerva Math: No License
    \item OlympiadBench: Apache-2.0 License
    \item DAPO-Math-17k: Apache-2.0 License
    \item DeepMath-103k: MIT License
    \item Qwen3 Models: Apache-2.0 License
    \item ROLL: Apache-2.0 License
    \item vLLM: Apache-2.0 License
    \item Megatron: Apache-2.0 License
\end{itemize}

%%%%%%%%%%%%%%%%%%%%%%%%%%%%%%%%%%%%%%%%%%%%%%%%%%%%%%%%%%%%

\end{document}